\RequirePackage{fix-cm}
\documentclass[smallextended]{svjour3}       
\smartqed  
\usepackage{graphicx}
\usepackage{mathptmx}         
\usepackage{siunitx}
\usepackage{booktabs}
\usepackage{etoolbox}
\usepackage{cite}
\usepackage{multirow}
\usepackage{hhline}

\usepackage{xcolor}
\usepackage{hyperref}
\hypersetup{
    colorlinks=true,
    linkcolor=blue,
    filecolor=magenta,      
    urlcolor=cyan,
    citecolor=darkgray,
    breaklinks=true,
    pdftitle={Complexity and Aesthetics in Generative and Evolutionary Ar},
    pdfpagemode=FullScreen,
}

\graphicspath{{Figures/}}

\usepackage{latexsym}
 \journalname{Genetic Programming and Evolvable Machines}
\begin{document}

\title{Complexity and Aesthetics in Generative and Evolutionary Art
}


\author{Jon McCormack         \and
        Camilo Cruz Gambardella 
}


\institute{Jon McCormack \at
              SensiLab, Monash University \\
              Melbourne, Australia\\
              \email{Jon.McCormack@monash.edu}           
           \and
           Camilo Cruz Gambardella \at
              SensiLab, Monash University \\
              Melbourne, Australia\\
              \email{Camilo.CruzGambardella@monash.edu}     
}

\date{Received: 30 May 2021 / Accepted: date}

\maketitle

\begin{abstract}
In this paper we examine the concept of complexity as it applies to generative and evolutionary art and design. Complexity has many different, discipline specific definitions, such as complexity in physical systems (entropy), algorithmic measures of information complexity and the field of ``complex systems''. We apply a series of different complexity measures to three different evolutionary art datasets and look at the correlations between complexity and individual aesthetic judgement by the artist (in the case of two datasets) or the physically measured complexity of generative 3D forms. Our results show that the degree of correlation is different for each set and measure, indicating that there is no overall ``better'' measure. However, specific measures do perform well on individual datasets, indicating that careful choice can increase the value of using such measures. We then assess the value of complexity measures for the audience by undertaking a large-scale survey on the perception of complexity and aesthetics. We conclude by discussing the value of direct measures in generative and evolutionary art,  reinforcing recent findings from neuroimaging and psychology which suggest human aesthetic judgement is informed by many extrinsic factors beyond the measurable properties of the object being judged.  

\end{abstract}

%
\section{Introduction}
\label{s:introduction}

Complexity is a topic of endless fascination in both art and science. For hundreds of years scholars, philosophers and artists have sought to understand what it means for something to be ``complex'' and why we are drawn to complex phenomena and visual complexity. Today, there are many different understandings of complexity, from information theory, physics, psychology, neuroscience and aesthetic theory \cite{Prigogine1980,Crutchfield1994b,GellMann1995,Wolfram2002, forsythe2011predicting}.

In this paper we revisit the concept of complexity, with a view to understanding if it can be useful for the generative or evolutionary artist. The application of complexity measures and their relation to aesthetics in generative and evolutionary art are numerous (see e.g.~\cite{Johnson2019} for an overview). A number of researchers have tested complexity measures as candidates for fitness measures in evolutionary art systems for example. We are interested in the value of complexity for both the individual \emph{artist or designer}, and the audiences experiencing their work. Put another way, we are asking what complexity can tell us about an individual artist's personal aesthetic taste or judgement, and how individual notions of complexity and aesthetics differ from general audience perceptions. 

A long held intuition is that visual aesthetics are related to an artefact's order and complexity \cite{berlyne1971aesthetics, Klinger2000, machado2015computerized}.
From a human perspective, complexity is often regarded as the amount of ``processing effort'' required to make sense of an artefact. Too complex and the form becomes unreadable, too ordered and one quickly looses interest.
Birkhoff famously formalised an aesthetic measure $M=O/C$, the ratio of order to complexity \cite{Birkhoff1933}, and similar approaches have built on this idea. Berlyne and colleagues defined visual complexity as ``irregularities in the spatial elements'' that compose a form \cite{papadimitriou2020spatial}, which lead to the formalisation of the relationship between pleasantness and complexity as an ``inverted-U'' \cite{berlyne1971aesthetics}. That is, by increasing the complexity of an artefact beyond the ``optimum'' value for aesthetic preference, its appeal starts to decline \cite{sun2014relationship}. However, researchers have noted the ``poor predictive validity'' of Berlyne's model \cite{forsythe2011predicting}. Another example is Biderman's theory of ``geons'', which proposes that human understanding of spatial objects depends on how discernible their basic geometric components are \cite{Biederman1993, papadimitriou2020spatial} Thus, the harder an object is to decompose into primary elements, the more complex we perceive it is. This is the basis for some image compression techniques, which are also often used as a measure of visual complexity \cite{Lakhal2020}. 

More recent surveys and analysis of computational aesthetics trace the history \cite{Greenfield2005, Hoenig2005} and current state of research in this area \cite{Johnson2019}. Some approaches introduce symmetry as a counterbalance to complexity, situating aesthetic appeal somewhere within the range spanning between these two properties \cite{papadimitriou2020spatial}. The most recent approaches combine measures of algorithmic complexity with different forms of filtering or processing to eliminate visual noise but retain overall detail \cite{Zanette2018, Lakhal2020}.

Multiple attempts to craft automated methods for the aesthetic judgement of images have made use of complexity measures. Moreover, some of these show encouraging results. In this paper we test a broad selection of these methods on three different image datasets produced using generative evolutionary art systems. All of the images in these datasets have their own ``aesthetic'' score as a basis for understanding the aesthetic judgements of the system's creator. We then extend our analysis from the relationship between complexity measures and the individual artist to that of human perception more generally. Using a public survey of people's judgements on the complexity and aesthetics of the same three datasets, we analyse the relationship between general perceptions and complexity measures. We also analyse the relationship between individual artist evaluations and more general human perception of complexity and aesthetics.

\section{Complexity and Aesthetic Measure}

Computational methods used to calculate image complexity are based on the definitions of complexity described the previous section (Section \ref{s:introduction}): i.e.~the amount of ``effort'' required to reproduce the contents of the image, as well as the way in which the patterns found in an image can be decomposed. Some methods have been proposed as useful measures of aesthetic appeal, or for predicting a viewer's preference for specific kinds of images.
In this section we outline those relevant for our research.

In the late 1990s, Machado and Cardoso proposed a method to determine aesthetic value of images derived from their interpretation of the process that humans follow when experiencing an aesthetic artefact \cite{Machado98}. Their method used the ratio of \emph{Image Complexity} -- a proxy for the complexity of the art itself -- to \emph{Processing Complexity} -- a proxy of the process humans use to make sense of an image --  as an approximation of how humans perceive images. 

In 2010, den Heijer and Eiben compared four different aesthetic measures on a simple evolutionary art system \cite{den_Heijer_2010}, including Machado and Cardoso's \emph{Image Complexity} / \emph{Processing Complexity} ratio, Ross \& Ralph's colour gradient bell curve, and the fractal dimension of the image. Their experiments demonstrated that, when used as fitness functions, different metrics yielded stylistically different results, indicating that each assessment method biases the particular image features or properties being evaluated. Interestingly, when interchanged -- when the results evolved with one metric are evaluated with another -- metrics showed different affinities, suggesting that regardless of the specificity of each individual measure, there are some commonalities between them.

A study by Forsythe and colleagues looked at the relationship between perceptions of complexity and computable measures \cite{forsythe2011predicting}. Using a database of 800 images composed of abstract artistic, abstract decorative, figurative representational, figurative decorative and environmental scene photographs, participants ($N=240$) were shown stratified samples for 5 seconds on a computer display and asked to rate each image's complexity (``the amount of detail or intricacy'') on a Likert scale. The results showed ``significant correlation'' between image compression and perceived complexity ($r_s = 0.51, p < 0.01$ for \emph{jpeg} compression and abstract artistic images, for example). The researchers also noted that familiarity with an image reduces its perceived complexity. The study also looked at the relationship between perceptions of beauty and complexity, and complexity and fractal dimension, finding ``fractal dimension alone cannot account for judgements of beauty''.

\section{Experiments}
\label{s:experimets}
To try and answer our question about the role and value of complexity measures in developing generative or evolutionary art systems, we compared a variety of complexity measures on three different generative art datasets, evaluating them for correlation with human or physical measures of aesthetics and complexity.

\subsection{Complexity Measures}
\label{ss:complexity-measures}
We tested a number of different complexity measures described in the literature to see how they correlated with individual evaluations of aesthetics. We first briefly introduce each measure here and go into more detail on specific measures later in the paper.

\begin{description}
    \item[\textit{Entropy} ($S$):]{the image data entropy measured using the luminance histogram (base $e$).}
    \item[\textit{Energy} ($E$):]{the data energy of the image.}
    \item[\textit{Contours} ($T$):]{the  number  of  lines  required  to  describe component boundaries detected in the image. The image first undergoes a morphological binarisation (reduction to a binary image that differentiates component boundaries) before detecting the boundaries.}
    \item[\textit{Euler} ($\gamma$):]{the morphological Euler number of the image (effectively a count of the number of connected regions minus the number of  holes). As with the $T$ measure, the image is first transformed using a morphological binarisation.}
    \item[\textit{Algorithmic Complexity} ($C_a$):]{measure of the algorithmic complexity of the image using the method described in \cite{Lakhal2020}.} Effectively the compression ratio of the image using Lempil-Ziv-Welch lossless compression.
    \item[\textit{Structural Complexity} ($C_s$):]{measure of the structural complexity, or ``noiseless entropy'' of an image using the method described in \cite{Lakhal2020}.}
    \item[\textit{Machado-Cardoso Complexity} ($C_{mc}$):]{a complexity measure used in \cite{machado2015}, without edge detection pre-processing.}
    \item[\textit{Machado-Cardoso Complexity with edge detection} ($C_{mc}^E$):]{the $C_{mc}$ measure with pre-processing of the image using a Sobel edge detection filter.}
    \item[\textit{Fractal Dimension} ($D$):]{fractal dimension of the image calculated using the box-counting method \cite{forsythe2011predicting}.}
    \item[\textit{Fractal Aesthetic} ($D_a$)]{aesthetic measure similar to that used in \cite{den_Heijer_2010}, based on the fractal dimension of the image fitted to a Gaussian curve with peak at 1.35. This value is chosen based on an empirical studies of aesthetic preference for fractal dimension \cite{Taylor2001}.}
\end{description}

While each of these measures is in some sense concerned with measuring image complexity, each measure's underlying basis is different. \emph{Entropy} ($S$) and \emph{Energy} ($E$) measures are based on information theoretic understandings of complexity but concern only the distribution of intensity, while \emph{Contours} ($T$) and \emph{Euler} ($\gamma$) try to directly count the number of lines or features in the image, somewhat in line with perceptual notions of complexity \cite{Eysenck1968}. Lakhal et.~al's \emph{Algorithmic Complexity} ($C_a$) and \emph{Machado \& Cardoso's Complexity} ($C_{mc}$) measures use algorithmic or Kolmogrov-like understandings of complexity, relying on image compression algorithms to proxy for visual complexity. Lakhal et.~al also define a \emph{Structural Complexity} measure ($C_s$) designed to address the limitations of algorithmic complexity measures in relation to high frequency noise or many fine details. This is achieved by a series of ``coarse-graining'' operations, effectively low-pass filtering the image to remove high frequency detail along with the quantisation of intensity.  Finally, the fractal methods recognise self-similar features as proxies for complexity. They are based on past analysis of art images that claim to have demonstrated relationships between fractal dimension and aesthetics \cite{Peitgen1986,Taylor:1999,forsythe2011predicting}.

\subsection{Datasets}
\label{ss:datasets}

\begin{figure}
\begin{center}
  \begin{tabular}{ccc}
\includegraphics[width=0.3\textwidth]{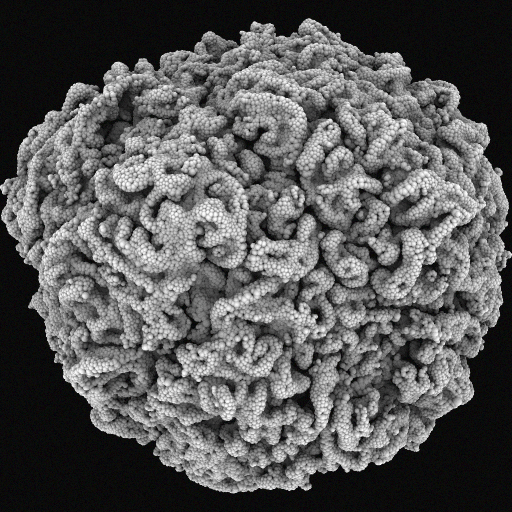} & \includegraphics[width=0.3\textwidth]{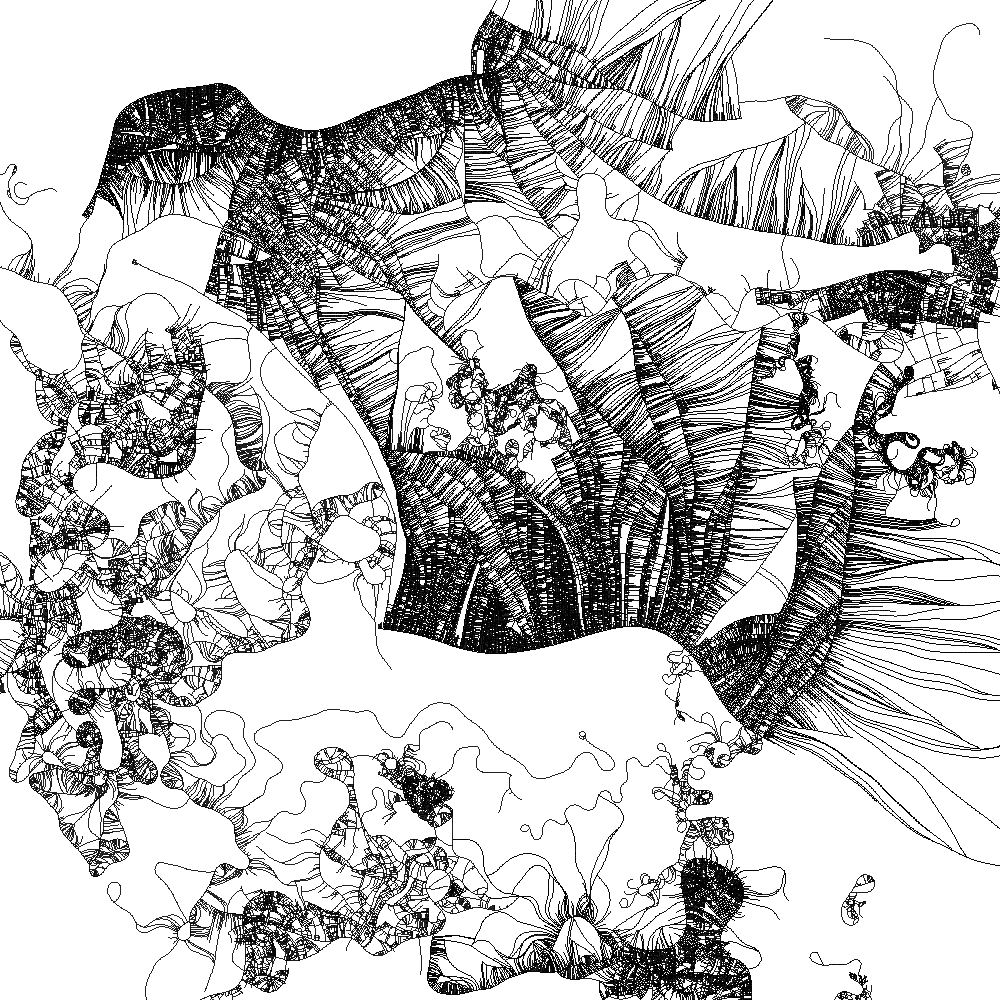} & \includegraphics[width=0.3\textwidth]{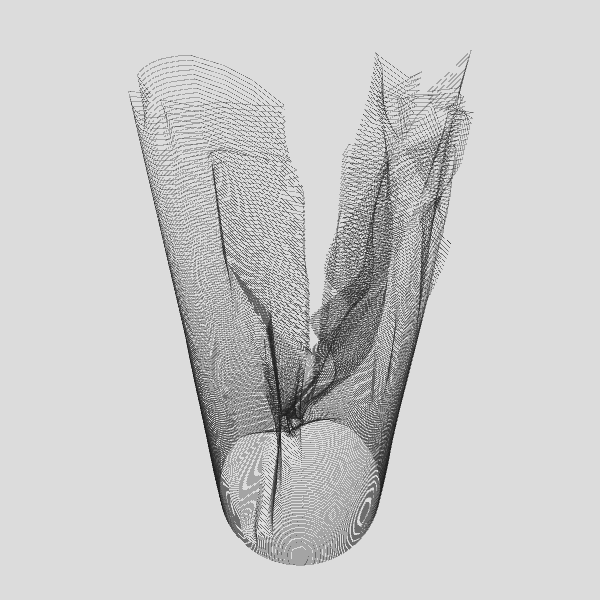} \\
a & b & c
  \end{tabular}
\end{center}
\caption{Example images from the Lomas (a), Line Drawings (b) and 3D DLA Forms (c) datasets.} \label{f:datasets}
\end{figure}

For the experiments described in this paper, we worked with three different generative art datasets (Figure \ref{f:datasets}). As the goal of this work was to understand the effectiveness of complexity measures in actual generative art applications, we wanted to work with artistic systems of demonstrated success, rather than invented or ``toy'' systems often used for validating research. This allows us to understand the ecological validity \cite{Brunswik1956} of any system or technique developed. Ecological validity requires the assessment of creative systems in the typical environments and contexts under which they are actually developed and used, as opposed to laboratory or artificially constructed settings. It is considered an important methodology for validating research in the creative and performing arts \cite{Jausovec2011}. Additionally, all the datasets are open access, allowing others to validate new methods on the same data.

\subsubsection{Dataset 1: Andy Lomas' Morphogenetic Forms}
\label{sss:lomasDataSet}
This dataset \cite{LomasDS2020} consists of 1,774 images generated using a 3D morphogenetic form generation system, developed by computer artist Andy Lomas \cite{Lomas2016,Lomas2018}. Each image is a two-dimensional rendering ($512 \times 512$ pixels) of a three-dimensional form that has been algorithmically ``grown'' from 12 numeric parameters. The images were evolved using an \emph{Interactive Genetic Algorithm} (IGA)-like approach with the software \emph{Species Explorer} \cite{Lomas2016,Lomas2018}. As the 2D images, not the raw 3D models are evaluated by the artist, we perform our analysis similarly.

The dataset contains an integer numeric aesthetic rating score for each form (ranging from 0 to 10, with 1 the lowest and 10 the highest, 0 meaning a failure case where the generative system terminated without generating a form or the result was not rated). These ratings were all performed by Lomas, so represent his personal aesthetic preferences. Additionally, each form is categorised by Lomas into one of eight distinct categories (these categories were not used in the experiments described in this paper).

\subsubsection{Dataset 2: DLA 3D Prints}
\label{sss:dla-3d-prints}

\begin{figure}
\begin{centering}
\includegraphics[width=0.65\textwidth]{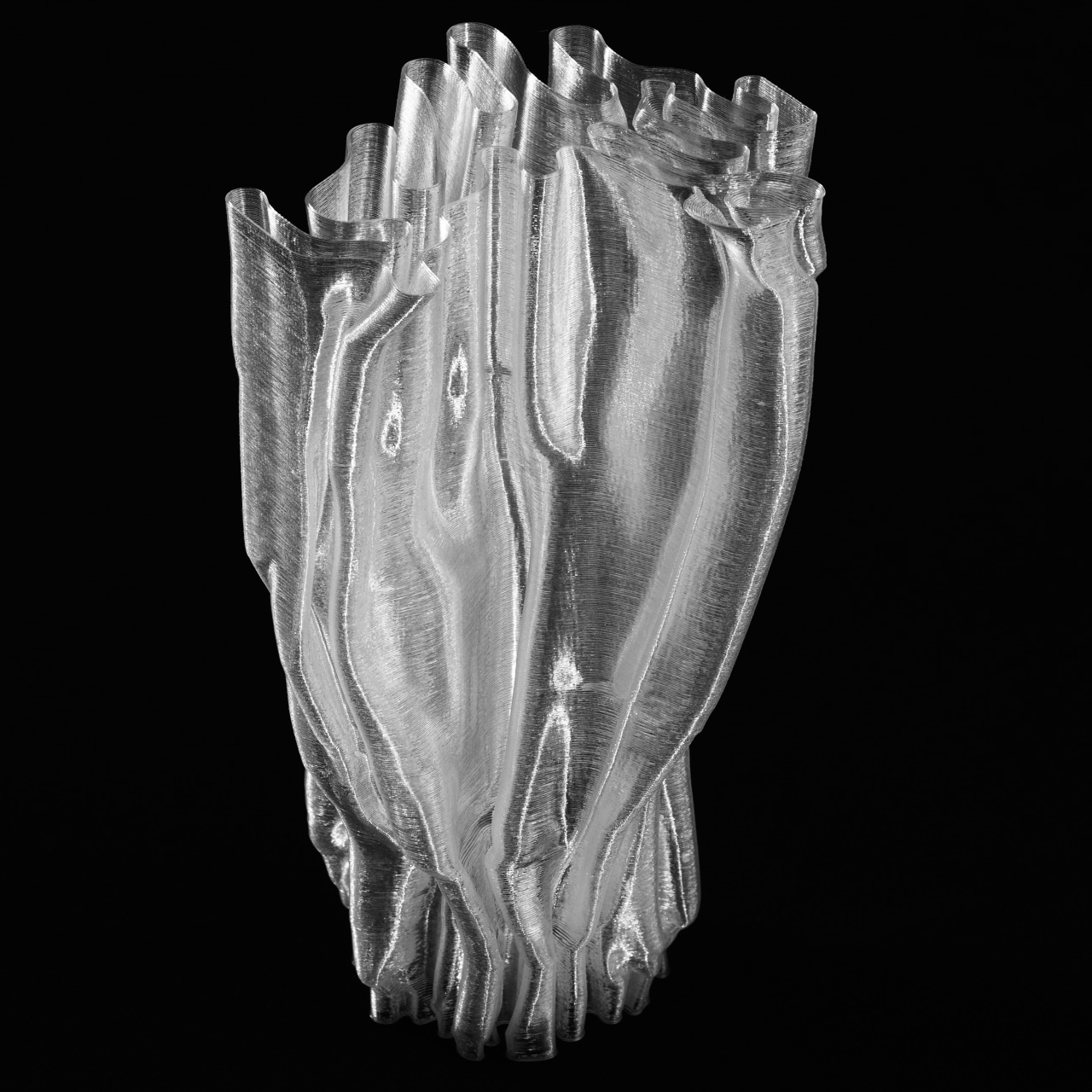}
\caption{Example 3D printed form from the DLA 3D Prints dataset.}
\label{f:heroPrint}
\end{centering}
\end{figure}

This dataset \cite{McCormack2021_DLADataset} consists of 2,500 3D forms created using a Differential Line Algorithm (DLA)-based method \cite{barlow1989differential, McCormack2021}. Multiple closed 2D line segments develop over time. At each time-step, the geometry is captured and forms a sequential z-layer in a 3D form. After several hundred time-steps, the final 3D form is generated, suitable for 3D printing (Figure \ref{f:heroPrint}).  Each image is $600 \times 600$ pixels resolution. Images in this set are 3D line renderings of the final form, from a perspective projection and orthographic projection in the $xy$ plane. In the experiments described here, we tested both the top-down orthographic images and perspective images, finding the perspective images gave better results and so are the ones reported here.

Rather than a artist-designated aesthetic measure, this dataset has a physically computed complexity measure. This measure is based on two geometric aspects of the 3D form:  \emph{convexity} (how much each layer deviates from its convex hull) and the quartile \emph{coefficient of dispersion} of angles between consecutive edges that make up each layer in the 3D form. These measures are calculated for each layer (weighted equally) and the final measure is the mean of all the layers in the form. This physical complexity measure appears to be a reasonable proxy to the visual complexity of the forms generated by the system.

\subsubsection{Dataset 3: Line Drawings}
\label{sss:line-drawings}
A set of 53 line drawings from a system designed by the first author, generated using an agent-based method based on the biological principles of niche construction \cite{McCormack2010,McCormackB09}. Each image is $1024 \times 1024$ pixels resolution. The dataset \cite{McCormack2021_ncDataset} also contains artist-assigned aesthetic scores normalised to the range $[0,1]$.

\subsection{Settings and Measure Details}
\label{ss:settings}

Our preliminary investigations showed that some measures are sensitive to parameter settings. The structural complexity measure ($C_s$) has two parameters: $r_{cg}$, a coarse-grain filter radius (in pixels), $\delta \in [0,0.5]$ a threshold for determining the black to white pixel ratio, $\eta \in [0,1]$ (white if $\eta \le \delta$, grey if $\delta < \eta \le 1 - \delta$, black for $\eta > 1 - \delta$). In the original study, the authors \cite{Lakhal2020} used values $(r_{cg},\delta) = (7,0.23)$ for one set of test images (abstract textures generated by Fourier synthesis) and $(13,0.12)$ for the second set (abstract random boxes placed using an inverse of the fractal box counting method) for $256 \times 256$ pixel resolution images. For the experiments described here, we used $(r_{cg},\delta) = (5,0.23)$ as our image sizes were larger and the images contain significant high frequency detail.

\begin{figure}
\begin{center}
  \begin{tabular}{ccc}
\includegraphics[width=0.3\textwidth]{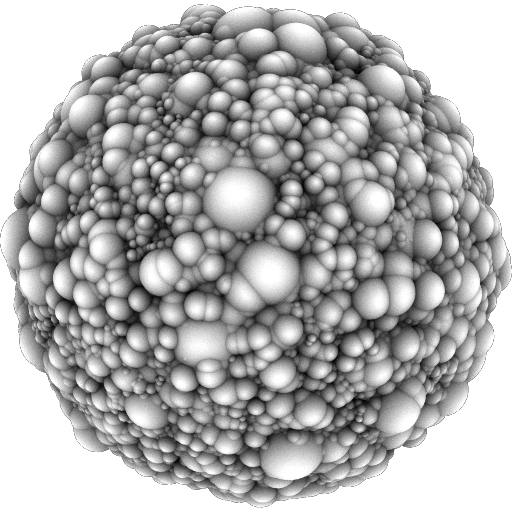} & \includegraphics[width=0.3\textwidth]{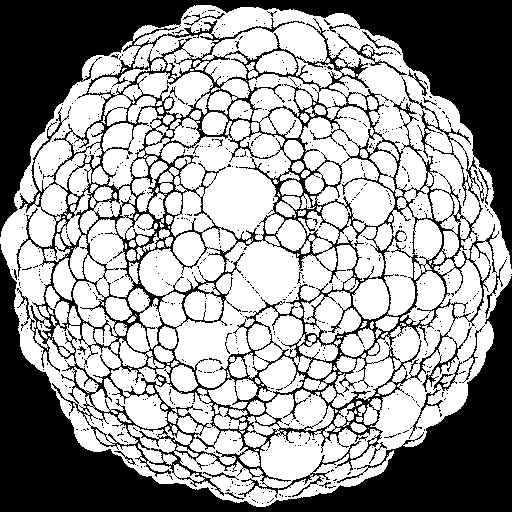} & \includegraphics[width=0.3\textwidth]{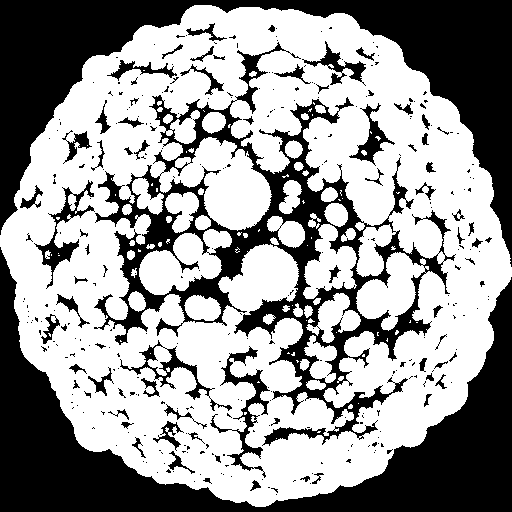} \\
Original Image & $r = 2$ & $r = 200$ \\
 & $D_a = 1.864$ & $D_a = 1.845$
  \end{tabular}
\end{center}
\caption{The effect of different adaptive binarisation radii on an image from the Lomas dataset. The fractal dimension ($D_a$) changes with different $r$ values.} \label{f:labin}
\end{figure}

For the fractal dimension measurements ($D,D_a$), images are pre-processed using a local adaptive binarisation process to convert the input image to a binary image (typically used to segment the foreground and background). A radius, $r$, is used to compute the local mean and standard deviation over $(2 r+1) \times (2 r+1)$ blocks centered on each pixel. Values above the mean of the $r$-range neighbourhood are replaced by 1, others by 0. Figure \ref{f:labin} shows a sample image from the Lomas dataset (left) with binary versions for $r = 2$ (middle) and $r = 200$ (right). Higher values of $r$ tend to reduce high frequency detail and result in a lower fractal dimension measurement. For the DLA 3D prints and Line Drawings datasets, which are already largely comprised of lines, the value of $r$ has negligible effect on the measurement.

Our Fractal Aesthetic Measure ($D_a$) is defined as:
\begin{equation}
    D_a(i) = exp(\frac{-(D(i) - p)^2}{2 \sigma^2}),
\end{equation}
where $p$ is the peak preference value for fractal dimension and $\sigma$ the width of the preference curve. $D_a$ returns a normalised aesthetic measure $\in [0,1]$. For the results reported here we used $(p,\sigma) = (1.35,0.2)$, based on prior findings for this preference \cite{Spehar2003}.

The Machado-Cardoso Complexity measure ($C_{mc})$) is defined as:
\begin{equation}
C_{mc}(i) = RMS(i, f(i)) \times \frac{s(f(i))}{s(i)},
\end{equation}
where $i$ is the input image, $RMS$ a function that returns the root mean squared error between its two arguments, $f$ a lossy encoding scheme for $i$ and $s$ a function that returns the size in bytes of its argument.\footnote{We adopted this measure as it specifically deals with complexity as defined in \cite{machado2015computerized}. Machado \& Cardoso also define an aesthetic measure as the ratio of image complexity to processing complexity \cite{Machado98}, as used by den Heijer \& Eiben in their comparison of aesthetic measures \cite{den_Heijer_2010}.} For the lossy encoding scheme, we used the standard JPEG image compression scheme with a compression level of 0.75 (0 is maximum compression).

\section{Results}
\label{s:results}
For each dataset we computed the full set of complexity measures (Section \ref{ss:complexity-measures}) on every image in the dataset, then computed the Pearson correlation coefficient between each measure and the human assigned aesthetic score (Lomas and Line Drawings datasets) or physically calculated complexity measure (DLA 3D Prints dataset).

\begin{table}
\caption{Lomas Datatset: Pearson's correlation coefficient values between image measurements and aesthetic score ($Sc$). The $C_{mc}$ complexity measure (bold) has the highest correlation with aesthetic score for this dataset. In all cases $p$-values are $< 1 \times 10^{-3}$}.
\label{tab:lomasCorrelations}
\robustify\bfseries
\begin{center}
\begin{tabular}{c|SSSSSSS[detect-weight]SSSS}
\toprule
 & {$S$} & {$E$} & {$T$} & {$\gamma$} & {$C_a$} & {$C_s$} & {$C_{mc}$} & {$C_{mc}^E$} & {$D$} & {$D_a$} & {Sc} \\
\midrule
$S$  &      1 &        &        &       &       &       &       &       &       &       & \\ 
$E$   & -0.989 &      1 &        &       &       &       &       &       &       &       & \\
$T$ &  0.425 & -0.375 &      1 &       &       &       &       &       &       &       & \\
$\gamma$    & -0.423 &  0.373 & -0.999 &     1 &       &       &       &       &       &       & \\
$C_a$    &  0.974 & -0.945 &  0.496 &-0.495 &     1 &       &       &       &       &       & \\
$C_s$    &  0.922 & -0.874 &  0.660 &-0.659 & 0.940 &     1 &       &       &       &       & \\
$C_{mc}$   &  0.793 & -0.732 &  0.590 &-0.589 & 0.907 & 0.860 &     1 &       &       &       & \\
$C_{mc}^E$ &  0.779 & -0.699 &  0.603 &-0.602 & 0.869 & 0.907 & 0.930 &     1 &       &       & \\
$D$     & -0.352 &  0.452 &  0.294 &-0.295 &-0.164 &-0.052 & 0.223 & 0.257 &    1  &       & \\
$D_a$    &  0.105 & -0.211 & -0.318 & 0.319 &-0.064 &-0.165 &-0.393 &-0.442 &-0.931 &     1 & \\
Sc    &  0.634 & -0.590 &  0.537 &-0.536 & 0.757 & 0.685 & \hspace{0.2cm} \bfseries{0.873} & 0.774 & 0.284 & -0.389&     1 \\
\bottomrule
\end{tabular}
\end{center}
\end{table}

\begin{table}
\caption{DLA 3D Prints Datatset: Pearson's correlation coefficient values between image measurements and physically computed complexity score ($Sc$). The $C_{s}$ structural complexity measure (bold) has the highest correlation with aesthetic score for this dataset.}
\label{tab:formsCorrelations}
\robustify\bfseries
\begin{center}
\begin{tabular}{c|SSSSSS[detect-weight]SSSSS}
\toprule
 & {$S$} & {$E$} & {$T$} & {$\gamma$} & {$C_a$} & {$C_s$} & {$C_{mc}$} & {$C_{mc}^E$} & {$D$} & {$D_a$} & {Sc} \\
\midrule
$S$  &      1 &        &        &       &       &       &       &       &       &       & \\ 
$E$   & -0.995 &      1 &        &       &       &       &       &       &       &       & \\
$T$ &  0.857 & -0.880 &      1 &       &       &       &       &       &       &       & \\
$\gamma$    & 0.107 &  -0.083 & -0.363 &     1 &       &       &       &       &       &       & \\
$C_a$    &  0.953 & -0.956 &  0.936 &-0.106 &     1 &       &       &       &       &       & \\
$C_s$    &  0.882 & -0.892 &  0.925 &-0.204 & 0.942 &     1 &       &       &       &       & \\
$C_{mc}$   &  0.915 & -0.935 &  0.968 &-0.197 & 0.969 & 0.950 &     1 &       &       &       & \\
$C_{mc}^E$ &  0.914 & -0.935 &  0.961 &-0.188 & 0.965 & 0.954 & 0.999 &     1 &       &       & \\
$D$     & 0.928 &  -0.949 &  0.869 &0.012 &0.896 &0.801 & 0.898 & 0.895 &    1  &       & \\
$D_a$    &  -0.870 & -0.888 & -0.761 & -0.112 &-0.798 &-0.678 &-0.779 &-0.774 &-0.972 &     1 & \\
Sc    &  0.760 & -0.726 &  0.652 &-0.066 & 0.756 & \hspace{0.15cm} \bfseries{ 0.774} & 0.704 & 0.706 & 0.550 & -0.434&     1 \\
\bottomrule
\end{tabular}
\end{center}
\end{table}

\begin{table}
\caption{Line Drawings Datatset: Pearson's correlation coefficient values between image measurements and aesthetic score ($Sc$). The Contours $T$ measure (bold) has the highest correlation with aesthetic score for this dataset.}
\label{tab:ncCorrelations}
\robustify\bfseries
\begin{center}
\begin{tabular}{c|SSS[detect-weight]SSSSSSSS}
\toprule
 & {$S$} & {$E$} & {$T$} & {$\gamma$} & {$C_a$} & {$C_s$} & {$C_{mc}$} & {$C_{mc}^E$} & {$D$} & {$D_a$} & {Sc} \\
\midrule
$S$  &      1 &        &        &       &       &       &       &       &       &       & \\ 
$E$   & -0.910 & 1        &       &       &       &       &       &       &       & \\
$T$ & 0.558 & -0.677 &      1 &       &       &       &       &       &       &       & \\
$\gamma$    & -0.559 &  0.677 & -1.000 &     1 &       &       &       &       &       &       & \\
$C_a$    &  0.994 & -0.934 &  0.541 &-0.541 &     1 &       &       &       &       &       & \\
$C_s$    &  0.576 & -0.717 &  0.474 &-0.474 & 0.618 &     1 &       &       &       &       & \\
$C_{mc}$   &  0.515 & -0.690 &  0.233 &-0.233 & 0.592 & 0.761 &     1 &       &       &       & \\
$C_{mc}^E$ &  0.648 & -0.811 &  0.312 &-0.312 & 0.712 & 0.822 & 0.927 &     1 &       &       & \\
$D$     & 0.580 &  -0.807 &  0.431 & -0.431 &0.640 & 0.835 & 0.867 & 0.914 &    1  &       & \\
$D_a$    &  -0.434 & 0.641 & -0.323 & 0.323 &-0.686 &-0.771 &-0.725 &-0.770 &-0.942 &     1 & \\
Sc    &  0.209 & -0.407 & \hspace{0.15cm} \bfseries{ 0.565} &-0.564 & 0.218 & 0.364 & 0.267 & 0.199 & 0.456 & -0.457 &     1 \\
\bottomrule
\end{tabular}
\end{center}
\end{table}

The results are shown for each dataset in Tables \ref{tab:lomasCorrelations} (Lomas), \ref{tab:formsCorrelations} (DLA 3D Prints) and \ref{tab:ncCorrelations} (Line Drawings) with the highest correlation measure shown in bold.

\begin{figure}
\begin{center}
  \begin{tabular}{cc}
\includegraphics[width=0.48\textwidth]{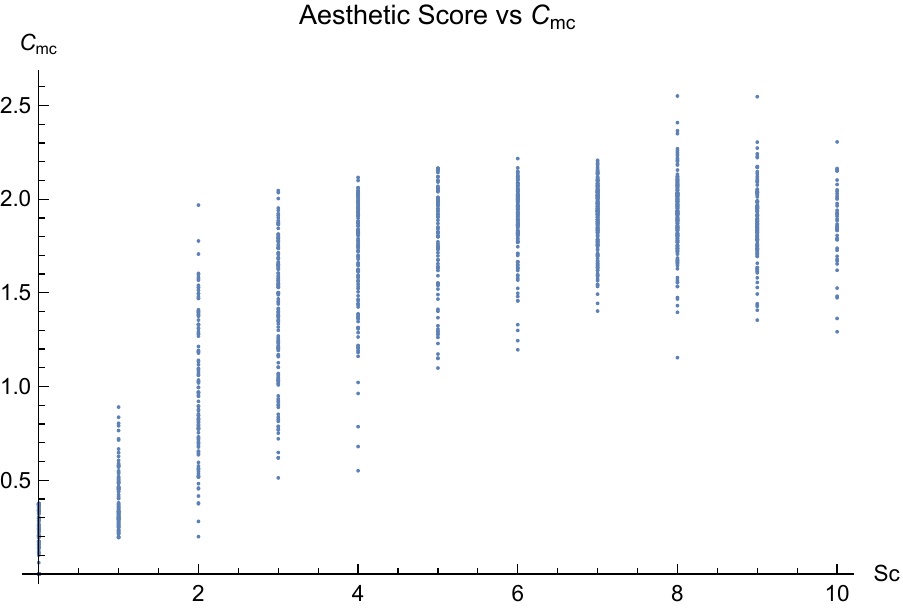} & \includegraphics[width=0.48\textwidth]{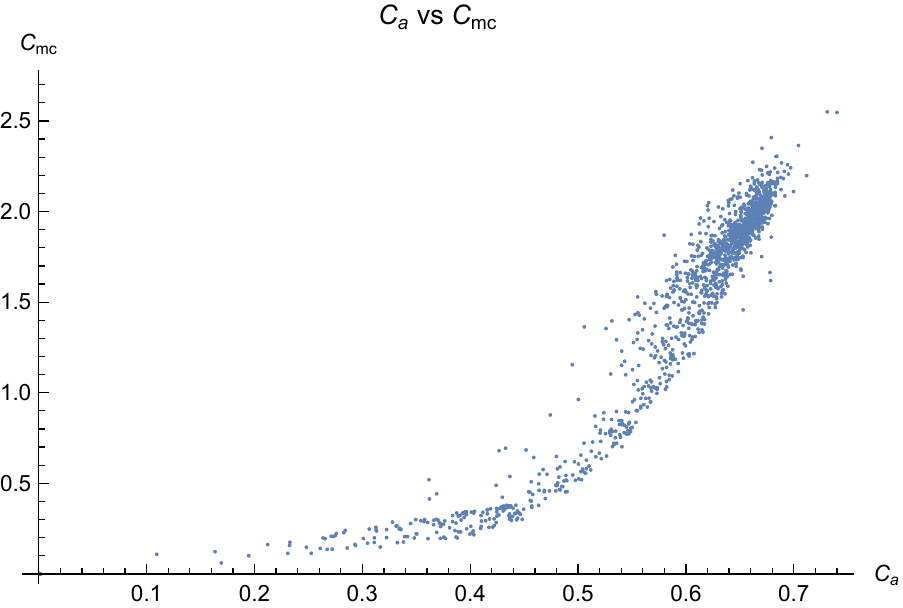}  \\
a & b
  \end{tabular}
\end{center}
\caption{The relationship between aesthetic score and $C_{mc}$ (a) and $C_a$ vs $C_{mc}$ (b) for the Lomas dataset. } \label{f:lomasPlots}
\end{figure}

The results show that a different complexity measure performed best for each dataset. For the \textbf{Lomas dataset} there is a strong correlation (0.873) between the artist assigned aesthetic score and the $C_{mc}$ complexity measure. Additionally, all the algorithmic and structural complexity measures are highly correlated. This is to be expected since they all involve image compression ratios and is further highlighted in Figure \ref{f:lomasPlots}, which shows a plot of aesthetic score vs $C_{mc}$ (a) and $C_s$ vs $C_{mc}$ (b). The banding in \ref{f:lomasPlots}a is due to the aesthetic scores being integers. A clear non-linear relationship between the complexity measures $C_s$ and $C_{mc}$ can be seen in \ref{f:lomasPlots}b.

Also of note is that fractal measures performed the worst of the measures tested. This seems to be confirmed visually: while certainly the images are complex (many are composed of 1 million or more cells) and have patterns at different scales, the patterns are not self-similar.

\begin{figure}
\begin{center}
  \begin{tabular}{cc}
\includegraphics[width=0.48\textwidth]{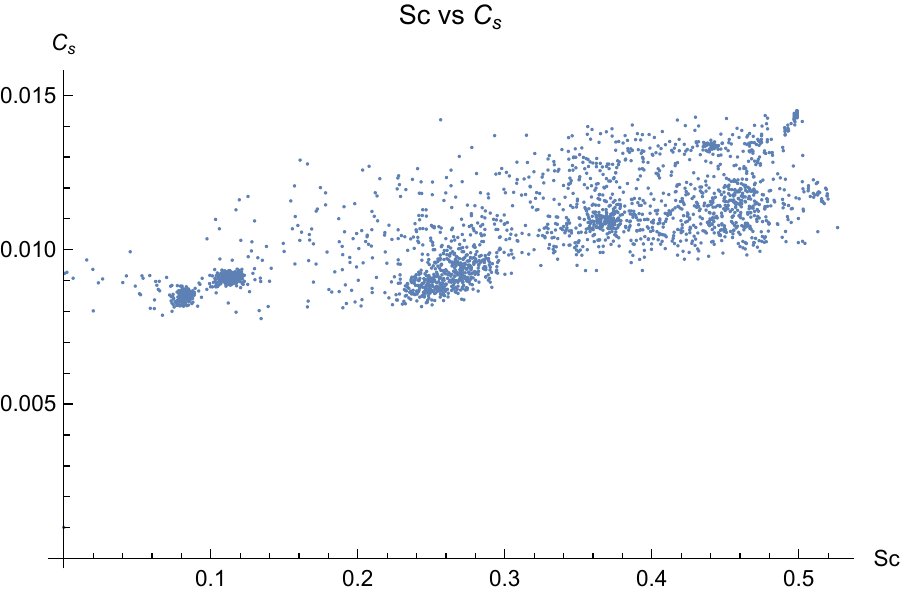} & \includegraphics[width=0.48\textwidth]{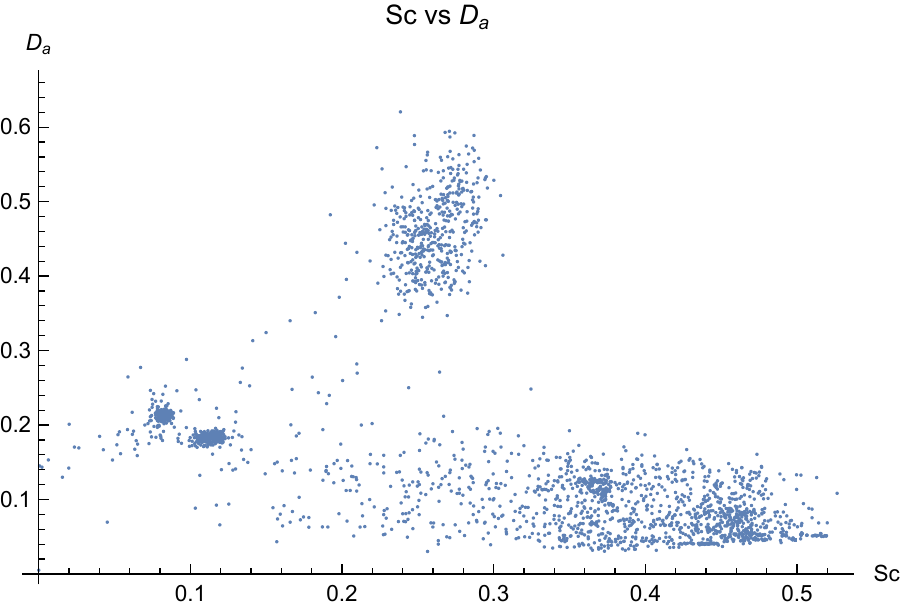}  \\
a & b
  \end{tabular}
\end{center}
\caption{The relationship between physical complexity score ($Sc$) and $C_{s}$ (a) and $D_a$ (b) for the DLA 3D Prints dataset. } \label{f:dlaPlots}
\end{figure}

For the \textbf{DLA 3D Prints} the most highly correlated measure was structural complexity ($C_s$) with a correlation of 0.774. The structural complexity aims to give a ``noiseless entropy'' measure by filtering high frequency spatial and intensity details. Given that the images are composed of many hundreds of thin lines stacked on top of each other, there is a significant amount of high frequency information, hence filtering is likely to give a better measure of real geometric details in each form. As can be seen in Figure \ref{f:dlaPlots}a, a clear correlation can be seen between the physical complexity ($Sc$) and Structural\footnote{Readers should not assume any direct relation between the terms ``structural'' and ``physical'' in relation to the complexity measures used here. Structural refers to image structures, whereas physical refers to characteristics of the 3D form's line segments.} Complexity measure ($C_s$).
Again we note that the fractal measures ($D,D_a$) had the lowest correlation and that all the algorithmic complexity measures are highly correlated. As shown in Figure \ref{f:dlaPlots}b however, there appears to be a kind of bifurcation and clustering in the relationship between $Sc$ and $D_a$, indicating a more complex, non-linear relationship between fractal dimension and complexity for this dataset.

\begin{figure}
\begin{center}
  \begin{tabular}{c|c}
\includegraphics[width=0.48\textwidth]{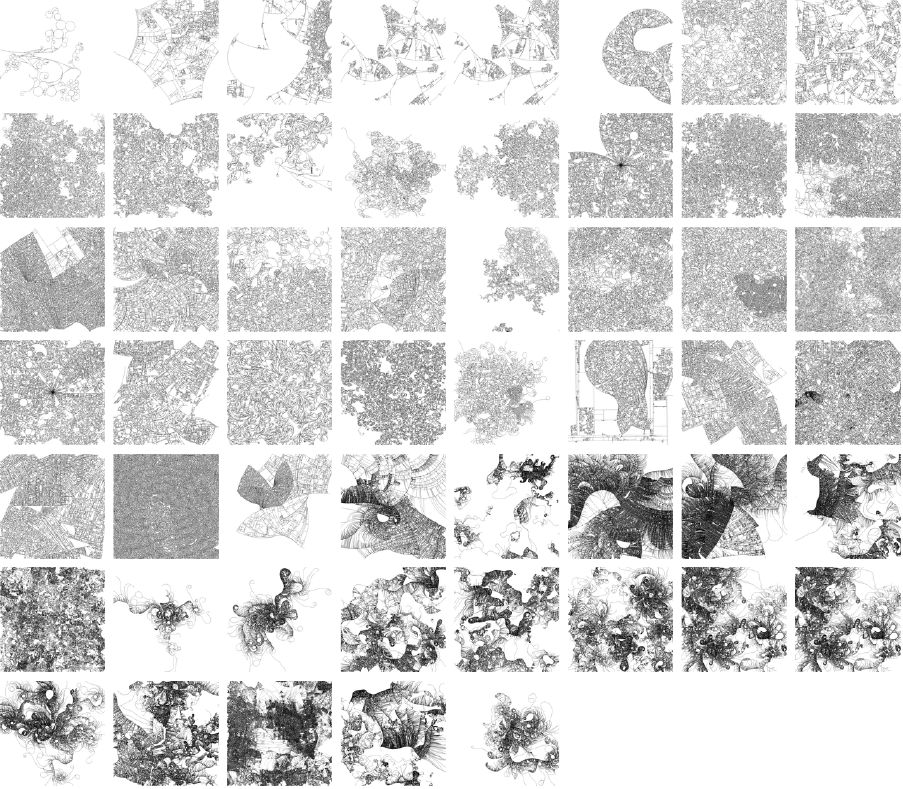} & \includegraphics[width=0.48\textwidth]{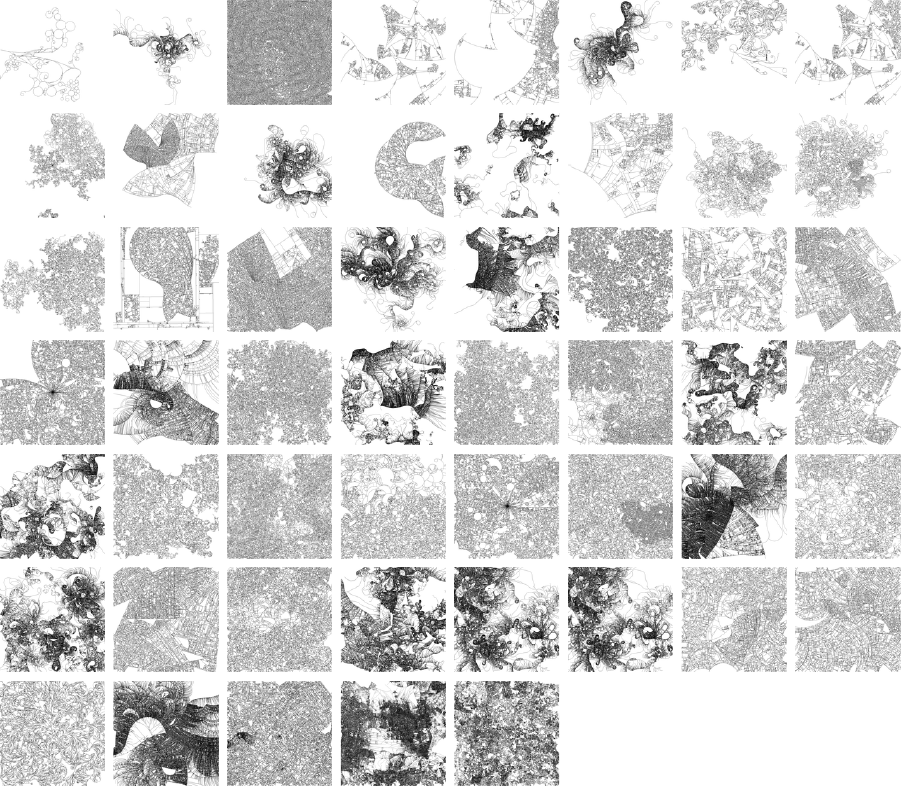}  \\
a & b
  \end{tabular}
\end{center}
\caption{Thumbnail grid of the entire Line Drawings dataset, ordered by (a) increasing aesthetic score (lowest top left, highest bottom right) and (b) by increasing structural complexity ($C_s$). As the size of the dataset is relatively small in comparison with the others, the images can be shown in the figure.} \label{f:linePlots}
\end{figure}

The \textbf{Line Drawings dataset} exhibited quite different results from the previous two. Here the Contours ($T$) measure had the highest, but only moderate, correlation with artist-assigned aesthetic scores (0.565). Given the nature of the drawings, measures designed to capture morphological structure seem most appropriate for this dataset. It is also interesting to note that the algorithmic complexity measures perform relatively poorly in this case. The original basis for the drawings came from the use of niche construction as a way to generate density variation in the images. The dataset contains images both with and without the use of niche construction, and generally those with niche construction are more highly ranked than those without. Figure \ref{f:linePlots} shows the entire dataset ordered in terms of artist-assigned aesthetic score (a) and structural complexity (b). The drawings with niche construction are easy to see as they are more highly ranked than those without. The structural complexity measure has greater difficulty in differentiating them (b).

With this in mind, we ran an additional image measure on this dataset that looks at asymmetry in intensity distribution (\emph{Skew}, $S_k$). Since the niche construction process results in contrasting areas of high and low density it was hypothesised that this measure might be able to better capture the differences. This measure had a correlation of 0.583 ($p = 4.5 \times 10^{-6})$, so better than any of the other measures, but not as high as the best complexity measures for the other two datasets.

%
%
\section{General Responses to Complexity and Aesthetics} 

Recall that so far, we have looked at the relationship between image complexity measures and the artist-assigned perceived aesthetic value (Lomas and Line Drawings datasets) or measured physical complexity (DLA 3D Prints dataset). These aesthetic or physical complexity evaluations are specific to the individual artist who designed the generative system or to the generative system itself.

To gain a better understanding of how the measures presented in the preceding sections relate to the perception of aesthetics and complexity generally by human observers, we developed a browser-based, online survey in which participants are prompted to express their preference when presented with image pairs selected from our three datasets. The survey was open to the general public and participants were invited via social media posts on the authors' personal accounts. Basic demographic information about participants -- age range, gender, experience with visual arts -- was collected. After 10 comparisons, participants were prompted to exit the survey or continue. Electing to continue allowed as many further comparisons as desired, participants could quit at any time by pressing the red ``Exit'' button, or by closing the browser window.

\begin{figure}
    \centering
    \begin{tabular}{ccc}
         \includegraphics[width=0.3\textwidth]{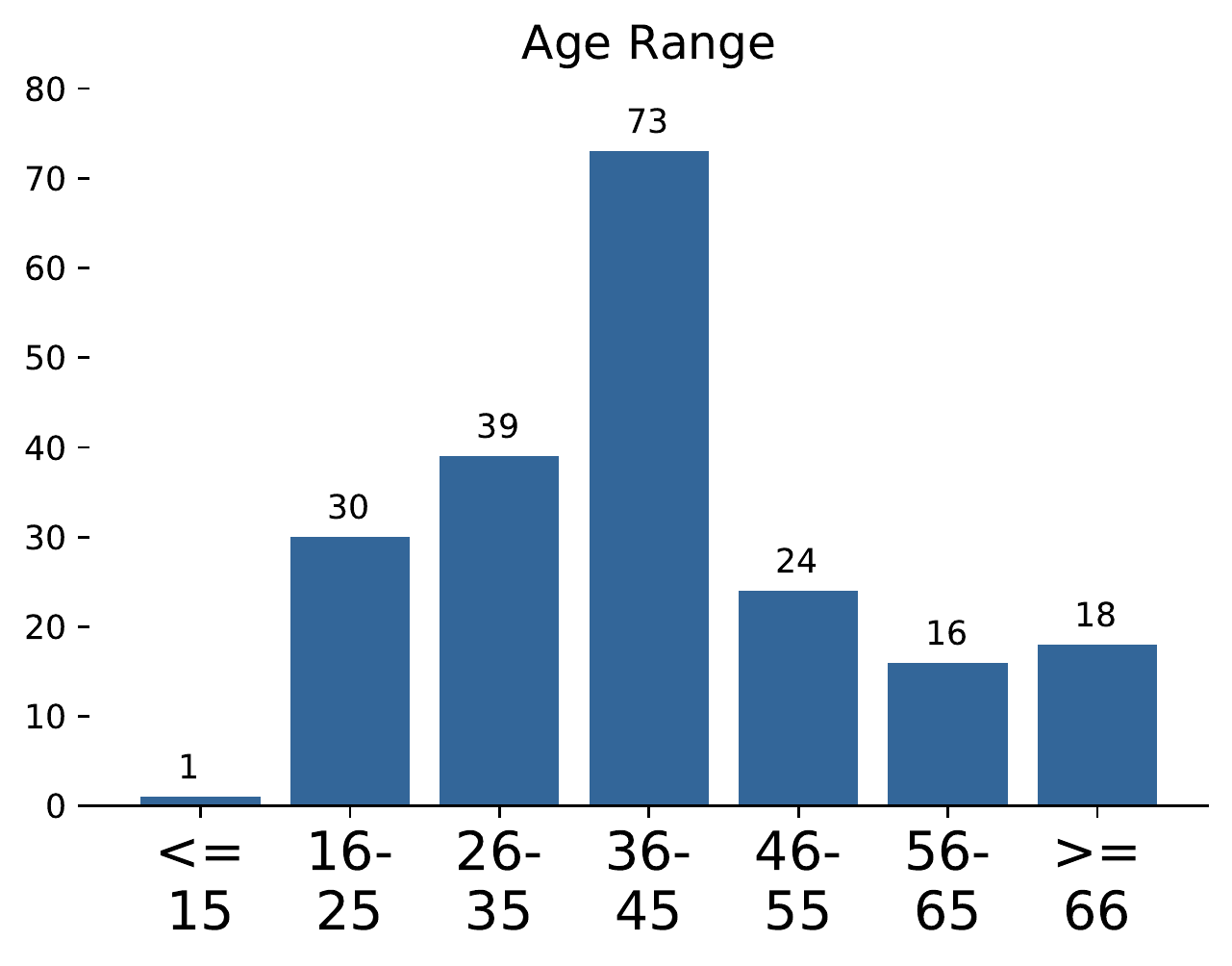} &
        \includegraphics[width=0.3\textwidth]{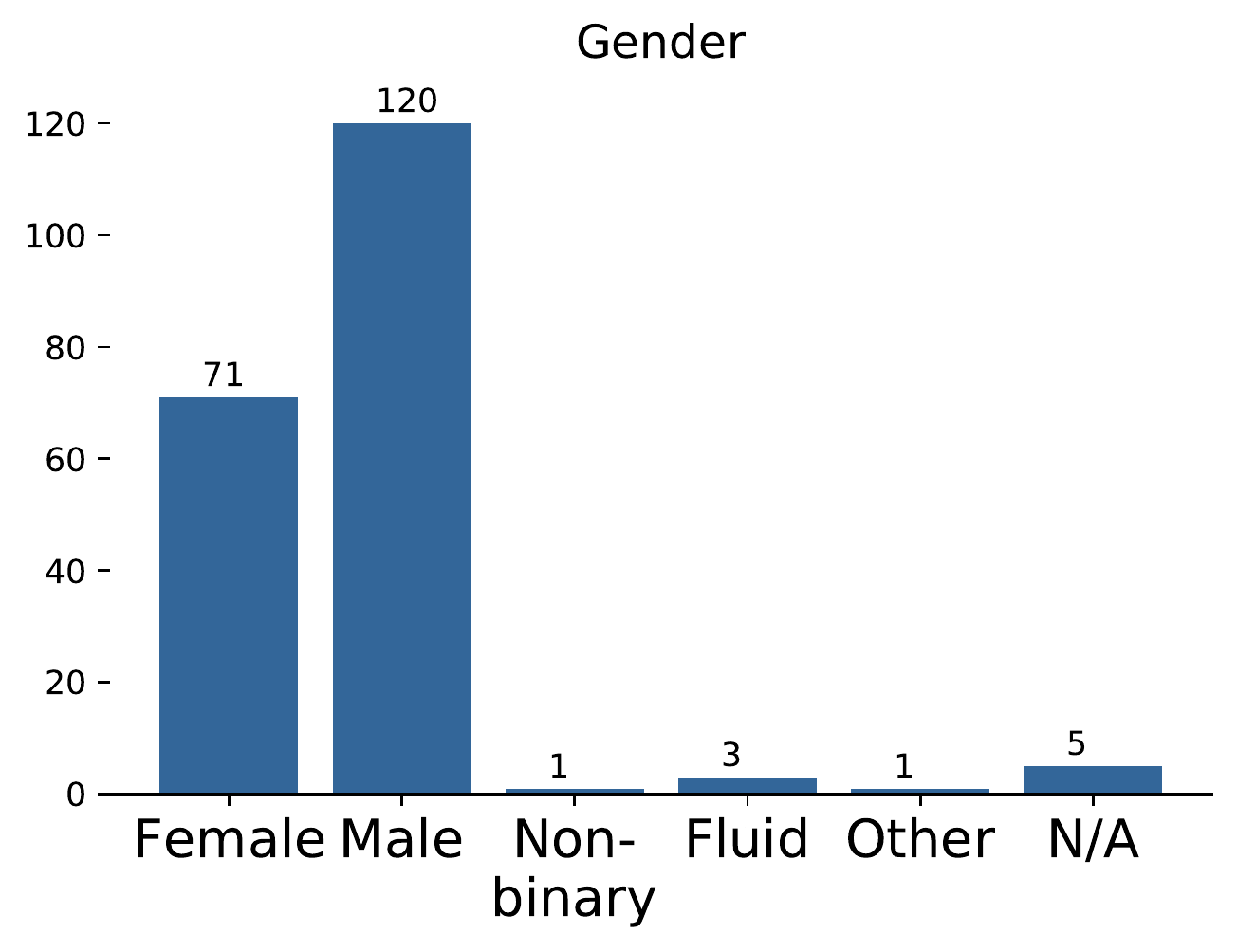} &
        \includegraphics[width=0.3\textwidth]{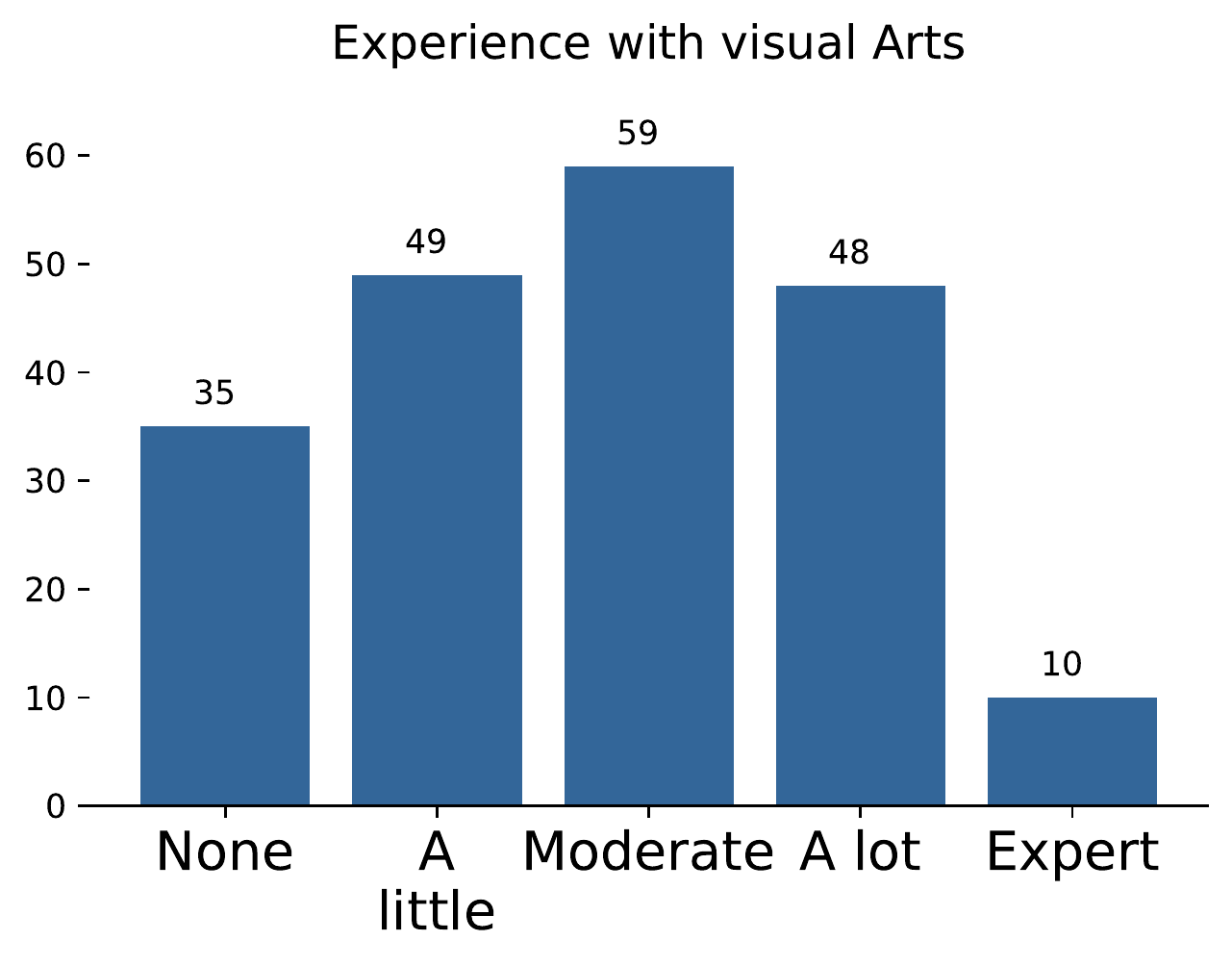}  \\
        a & b & c
    \end{tabular}

    \caption{Distributions of demographic information collected for the survey.}
    \label{f:demographics}
\end{figure}

At the time of reporting 201 respondents provided a total of 5,341 comparisons for the survey. The distribution of age-ranges, gender and expertise level are shown in Figure \ref{f:demographics}.

\subsection{Survey} \label{ss:survey}

\begin{figure}
    \centering
    \begin{tabular}{cc}
         \includegraphics[width=0.48\textwidth]{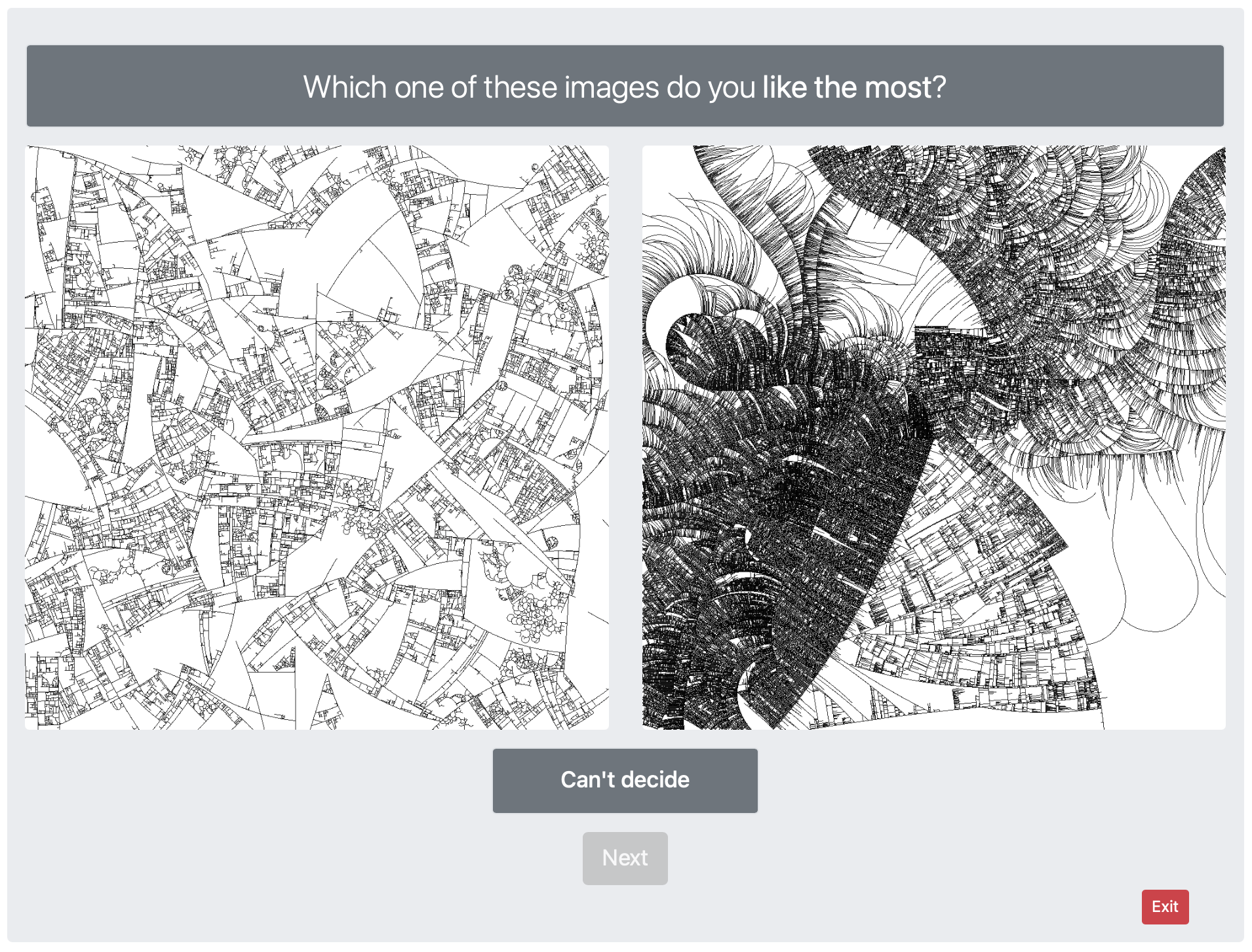} & \includegraphics[width=0.48\textwidth]{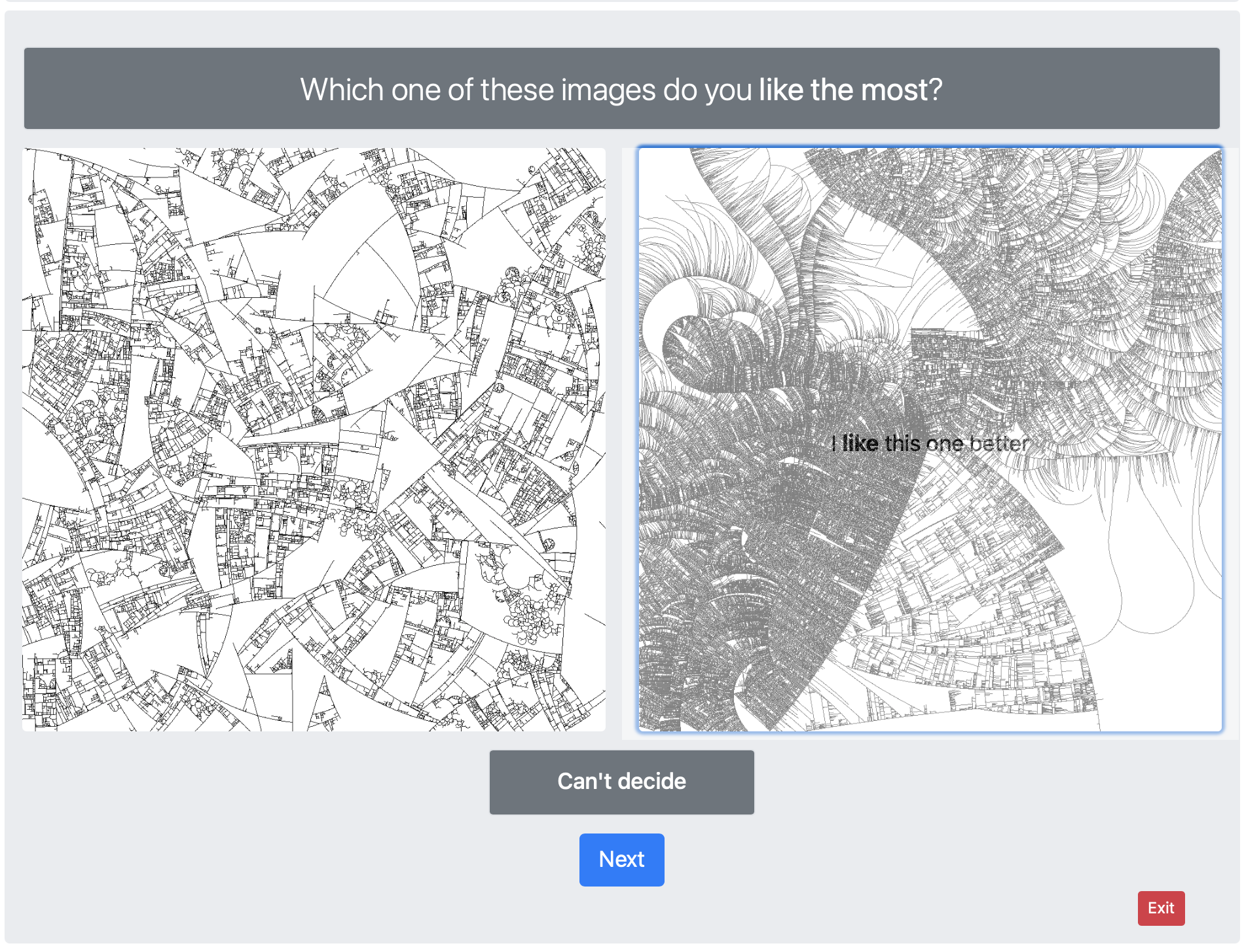}\\
         a & b
    \end{tabular}
    
    \caption{Screenshots of the survey question screen, in this example images from the Line Drawings dataset are displayed. a) Question as presented to participant. b) Question screen with option selected. A ``Can't decide'' option is provided and answers to survey questions have to be submitted using the `Next' button, giving participants the opportunity to review their selection.}
    \label{f:surveyMain}
\end{figure}

The goal of our survey was to evaluate participants' perception of complexity and aesthetic preference in relation to each of the three datasets. Pairwise comparisons were used, asking participants to compare two images from the same dataset and decide on a specific prompt by selecting one of the images. This method
provides participants with points of reference and eliminates the uncertainties associated with direct rating systems, such as Likert scales or judgements regarding the relative distance between items rated in isolation. Participants were presented with two randomly selected images from the same dataset and one of two possible comparison prompts: a) \emph{Which one of these images do you like the most?}, or b) \emph{Which of these images is more complex?}, which is also selected randomly (Figure \ref{f:surveyMain}a).

Visual cues were implemented in the user interface to help participants make a conscious decision. Firstly, when a prompt is presented to the user, the `Next' button is greyed out, indicating that a preference has to be entered. Then, once a selection has been made, the selected image is highlighted and a text overlay related to the prompt is displayed, as a way to  ensure that the answer being provided relates to the question (see Figure \ref{f:surveyMain}b for reference). Discarding any comparison duration over 5 minutes (only 16 comparisons), the median response time across all 5,325 comparisons was 7.7 seconds (mean: 10.4 seconds, sd: 10.8).

\subsection{Rating and ranking dataset entries}

The outcomes of the pairwise comparison method described in section \ref{ss:survey} were used to rank the images in our datasets. Rankings are determined using Glickman's ``Glicko System'' \cite{glickman1998glicko, glickman1999parameter}, originally developed to rate and rank chess players. The Glicko System incorporates a rating deviation ($RD$) measure that represents the reliability of a player's rating, calculated as a function of game outcomes and time between rating periods, in order to account for any changes that may have occurred while a player has not participated in any rated games (e.g. improving their ability via intensive training).

For image rankings, we considered pairwise comparisons as contests between images, where each image is a ``player'' competing to win against the other in the pair. The selected image is the winner of that contest (the \emph{can't decide} option represents a tie). To compute the rating of an image, we introduced two minor modifications to the standard Glicko System, tailoring the system to the nature of the data being processed. First, given that, unlike chess players, the images themselves do not improve or worsen over time, we disregard the time factor when calculating $RD$. The second modification is that we calculate rating and $RD$ after every ``match'' (comparison), instead of defining a rating period every $n$ matches. $RD$ still provides a measure of the confidence of a ranking, allowing us to filter those images with insufficient certainty regarding their ranking (for example too few comparisons or ``noise'' in the results). 

At the start of the survey all images were initialised with the same rating and $RD$ for their aesthetics ($A_a$) and complexity ($A_C$). We used $rating=1500$ and $RD=350$ (as recommended in \cite{glickman1998glicko}).

\subsection{Results}

We analysed the survey responses to see if there was any correlation between perceived complexity and aesthetics in the images. The results are summarised in table \ref{tab:surveyResults}. For the Lomas and DLA 3D Prints datasets, the survey results were first filtered, selecting only those images with an $RD < 290$ for both complexity and aesthetics to ensure sufficient confidence in the ranking. The Line Drawings dataset, being much smaller, had all images with an $RD < 290$. The sample size ($N$) after filtering (also expressed as percentage of the full dataset), along with the Pearson correlation ($r_s$) and $p$-value are shown for perceived complexity ($A_C$) and aesthetics ($A_a$) in the table.

\begin{table}
\begin{centering}
\caption{Correlation between the audience survey perceptions of complexity and aesthetics}
\label{tab:surveyResults}       
\begin{tabular}{lcrr}
\hline\noalign{\smallskip}
Dataset & $N$ (\%) &  $r_s(A_C,A_a)$ & $p$-value  \\
\noalign{\smallskip}\hline\noalign{\smallskip}
Lomas & 252 (14.4\%) & -0.140 & 0.026 \\
3D Forms & 141 (5.64\%) &  0.034 & 0.687 \\
Line Drawings & 52 (100\%) & -0.118 & 0.400 \\
\noalign{\smallskip}\hline
\end{tabular}
\end{centering}
\end{table}

\begin{figure}
\begin{center}
\begin{tabular}{ccc}
\includegraphics[width=0.29\textwidth]{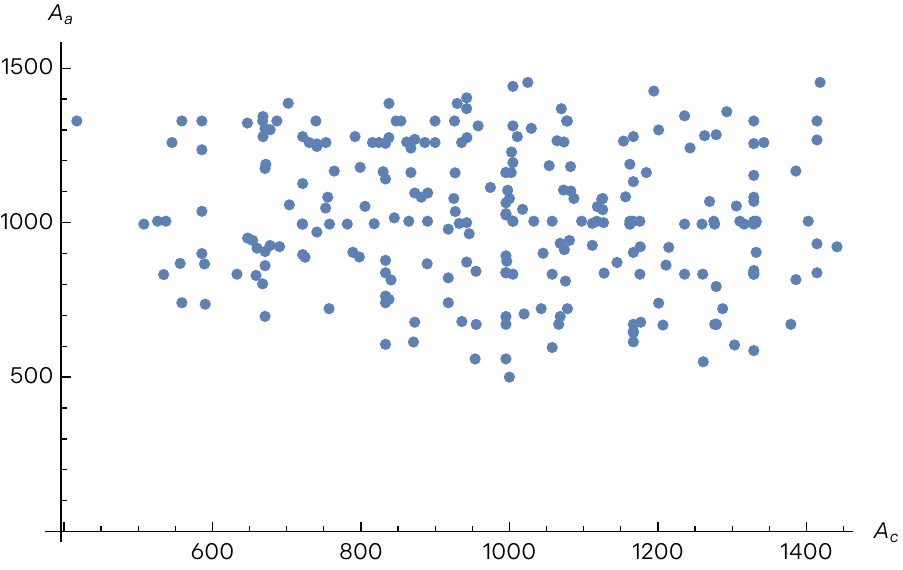} & \includegraphics[width=0.29\textwidth]{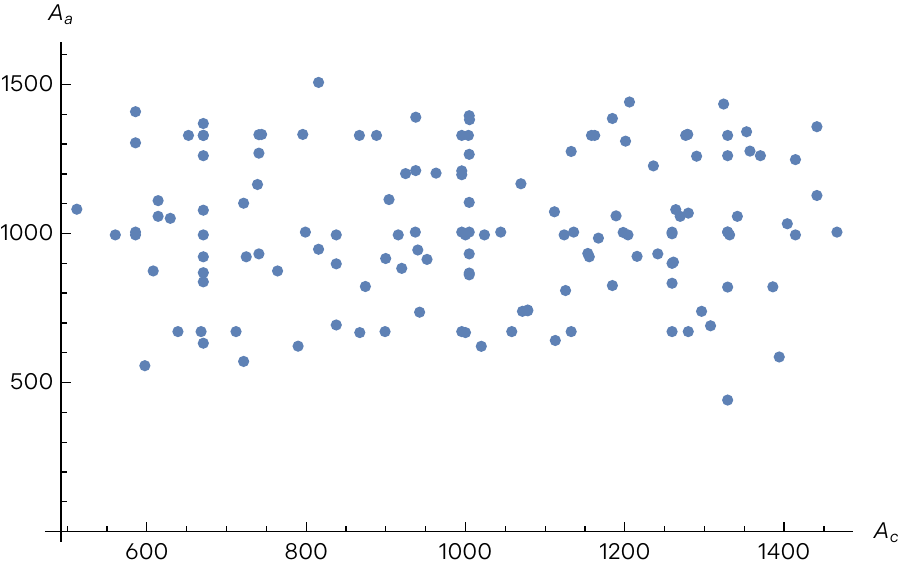} & \includegraphics[width=0.29\textwidth]{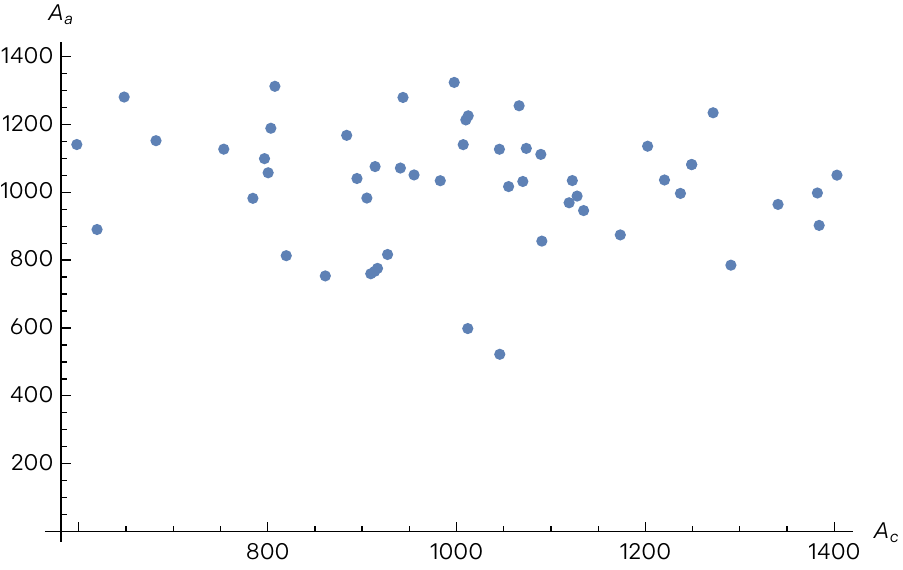} \\
Lomas & 3D Forms & Line Drawings
  \end{tabular}
\end{center}
\caption{Relationship between perceived complexity and aesthetics for each dataset} \label{f:CA-plots}
\end{figure}

As can be seen, the results do not confirm any significant correlation between the general perception of complexity and aesthetic value in an image. For the Lomas dataset, we can be reasonably confident that there is almost no correlation between perceived complexity and aesthetics. Figure \ref{f:CA-plots} shows the relationship between $A_C$ and $A_a$ for each dataset.

Given the lack of a discernible relationship between perceived complexity and aesthetics, we next analysed the data for relationships between the computational, individual artist-assigned, and survey measures. The results are summarised in table \ref{tab:surveyResults2}. The table shows the results for each dataset, first giving the Pearson score correlations between the artist assigned score (or complexity measure in the case of the DLA 3D Prints), the best complexity measure (from Section \ref{s:results}) and the survey results. The highest correlations between perceived complexity and aesthetics are shown next.

\begin{table}
\caption{Correlation between the audience survey perceptions of complexity and aesthetics with computational complexity measures and artist assigned scores}
\label{tab:surveyResults2}       
\begin{centering}
\begin{tabular}{lcccc}
\hline\noalign{\smallskip}
\multicolumn{4}{l}{\textit{Lomas Dataset}}\\
\noalign{\smallskip}
\multirow{ 2}{*}{Score Correlations} & & $C_{mc}$ & $A_C$ & $A_a$\\
 \cline{2-5}\noalign{\smallskip}
 & $Sc$ & 0.873 & 0.327 & 0.025 \\
\noalign{\smallskip}\hline\noalign{\smallskip}
\multirow{ 2}{*}{Best Correlations} & $A$ & $B$ & $r_s(A,B)$ & $p$-value \\
 \cline{2-5}\noalign{\smallskip}
 & $A_C$ & $C^{E}_{mc}$ & 0.407 & $<10^{-5}$ \\
  & $A_a$ & $D_a$ & 0.085 & 0.177 \\
\noalign{\smallskip}\hhline{=====}\noalign{\smallskip}
\multicolumn{4}{l}{\textit{DLA 3D Prints Dataset}}\\
\noalign{\smallskip}
\multirow{ 2}{*}{Score Correlations} & & $C_s$ & $A_C$ & $A_a$\\
 \cline{2-5}\noalign{\smallskip}
 & $Sc$ & 0.774 & 0.394 & 0.199 \\
\noalign{\smallskip}\hline\noalign{\smallskip}
\multirow{ 2}{*}{Best Correlations} & $A$ & $B$ & $r_s(A,B)$ & $p$-value \\
 \cline{2-5}\noalign{\smallskip}
 & $A_C$ & $C_s$ & 0.362 & $<10^{-4}$ \\
  & $A_a$ & $C_a$ & 0.226 & 0.007 \\
\noalign{\smallskip}\hhline{=====}\noalign{\smallskip}
\multicolumn{4}{l}{\textit{Line Drawings Dataset}}\\
\noalign{\smallskip}
\multirow{ 2}{*}{Score Correlations} & & $S_k$ & $A_C$ & $A_a$\\
 \cline{2-5}\noalign{\smallskip}
 & $Sc$ & 0.583 & 0.586 & 0.156 \\
\noalign{\smallskip}\hline\noalign{\smallskip}
\multirow{ 2}{*}{Best Correlations} & $A$ & $B$ & $r_s(A,B)$ & $p$-value \\
 \cline{2-5}\noalign{\smallskip}
 & $A_C$ & $S_k$ & 0.752 & $<10^{-5}$ \\
  & $A_a$ & $S$ & -0.467 & 0.0004 \\
\noalign{\smallskip}\hline
\end{tabular}
\end{centering}
\end{table}

The Lomas dataset had a high correlation between the the artist-assigned score ($Sc$) and the Machado-Cardoso Complexity measure ($C_{mc}$): 0.873. This was not reflected in the audience perceived complexity ($A_C$) or aesthetics ($A_a$) correlations, with only a low correlation with $A_C$ and no correlation with $A_a$. There is a mild correlation between audience perceptions of complexity with the Machado-Cardoso Complexity measure with edge processing ($C^{E}_{mc}$), which only just beat the other algorithmic complexity measures ($C_s$, $C_{mc}$, $C$), which were correlated in the 0.34-0.38 range. The Fractal Aesthetic measure had the highest correlation with perceived aesthetics, but the overall correlation is negligible for all the measures, suggesting no connection between audience perceived aesthetics and computational complexity measures.  

Similar results are observed for the DLA 3D Prints dataset. A strong correlation between the Structural Complexity measure ($C_s$) was less strongly reflected in the perceived complexity (0.327) and, again, no correlation between the score and perceived aesthetics. There were mild to weak correlations between the perceived complexity and aesthetics with the Structural Complexity ($C_s$) and Algorithmic Complexity ($C_a$) measures.

For the Line Drawings dataset, there was moderate correlation (0.583) between the artist-assigned score and Image Skew. A similar correlation was found with the audience perceptions of complexity (0.586), far less for aesthetics (0.156). The similarities between Image Skew and perceived complexity are reflected in the best correlations (0.752). Interestingly, an inverse correlation was the highest found between Image Entropy ($S$) and perceived aesthetics (-0.467).

%
%
\section{Discussion}
\label{s:discussion}

The results from Section \ref{s:results} show that there appears to be no single measure that is best to quantify image complexity in the the context of generative art. Hence it seems wise to select a measure most appropriate to the style or class of imagery or form being generated. In general, the image information measures ($C_{mc}$, $C_a$ and $C_s$) had the overall best results across all the datasets evaluated. 

It is also important to point out that, in general, computer synthesised imagery and in particular images generated by algorithmic methods, have important characteristics that differ from other images, such as photographs or paintings. Apart from any semantic differences or differentiation between figurative and abstract, intensity and spatial distributions in computer synthesised images differ from real world images. This is one reason why we selected datasets that are specific to the application of these measures (generative art and design), rather than human art datasets in general, for example. Our datasets are also exclusively greyscale images, eliminating colour as a consideration in our experiments.

In the second part of the paper, we compared the computational measures and individual artist-assigned scores with more general human perceptions of complexity and aesthetics, using a pairwise ranking survey to order each dataset in terms of perceived complexity and aesthetic preference, with 201 participants and over 5,300 comparisons. Based on the survey, we found no significant correlation between perceived complexity and perceived aesthetics for any of the datasets. General perceptions of complexity were at best, only mildly correlated with computational or artist-assigned measures. Aesthetic perceptions fared the worst, with no significant correlation between either computational measures or individual artist evaluations. In short, the computational measures tested cannot predict what an audience will like, but neither can the artist! (at least in the case of the datasets we used).

The rationale for this research was to further the question: \emph{how can complexity measures be usefully employed in generative and evolutionary art and design?} Based on the results presented in this paper, our answer is that -- if chosen appropriately -- they can be valuable aids for  coarse-level discrimination for an individual artist. Additionally, they are relatively quick to compute and work without prior training or exposure to large numbers of examples or training sets, as would be the case for neural network discriminators for example. Hence, complexity measures could be useful in filtering or ranking individuals in an IGA or used to help classify or select individuals for further enhancement using other methods. However they seem insufficient as fully autonomous fitness measures -- the human designer remains a vital and fundamental part of any good aesthetic evaluation.

\subsection{Aesthetic Judgement}
\label{ss:aestheticJudgement}

In Section \ref{s:introduction} we discussed possible relationships between complexity measures and aesthetics. It is worth reflecting further here on this relationship and the long-held ``open problem'' for evolutionary and generative art of quantifying aesthetic fitness \cite{McCormack2005a}. In contrast to studies that have looked at ``art'' images in general, we did not find a perceived correlation between complexity and aesthetics for the computer synthesised images in our datasets.

In the last decade or so, the biggest advances in the understanding of computational and human aesthetic judgements have come from (i) large, open access datasets of imagery with associated human aesthetic rankings and (ii) psychological and neuroscience discoveries on the mechanisms of forming an aesthetic judgement and what constitutes aesthetic experience.

In a recent paper, Skov summarised aesthetic appreciation from the perspective of neuroimaging \cite{Skov2019}. Some of the key findings included neuroscientific evidence suggesting that ``aesthetic appreciation is not a distinct neurobiological process assessing certain objects, but a general system, centered on the mesolimbic reward circuit, for assessing the hedonic value of any sensory object'' \cite{Skov2019}. Another important finding was that hedonic values are not solely determined by object properties. They are subject to numerous factors extrinsic to the object itself. Similar claims have come from psychological models \cite{Leder2014}. These findings suggest that any algorithmic measure of aesthetics that only considers an object's visual appearance ignores many other extrinsic factors that humans use to form an aesthetic judgement (including context, prior knowledge and experience, emotional state and affect). Hence they are unlikely to correlate strongly with human judgements generally. 

The results presented in this paper appear to tally with these findings. Complexity measures, carefully chosen for specific styles or types of generative art can capture some broad aspects of personal aesthetic judgement, but they are insufficient alone to fully replace human judgement and discretion. Using other techniques, such as deep learning, \emph{may} result in slightly better correlation to individual human judgement \cite{McCormackLomas2020b}, however such systems require training on large datsets which can be tedious and time-consuming for the artist and still do not do as well as the trained artist's eye in resolving aesthetic decisions.

While individual artists and designers may have a strong sense of complexity and aesthetics in their own systems, the survey results do not support the hypothesis that such a sense generalises. Prior studies have shown that perceptual qualities such as complexity and aesthetics change according to familiarity \cite{Forsythe2008}. A difference between artist assigned measures and perceptions generally is that artists work closely with their generative and evolutionary systems over long periods of time, meticulously studying thousands of similar images and becoming attuned to their nuances and differences. A less educated eye may not share such detailed discrimination as our survey results suggest.  

\section{Conclusion}
\label{s:conclusion}
Making and appreciating art is a shared human experience. Computers can expand and grow the creative possibilities available to artists and audiences. The fact that humans artists are successfully able to create and communicate artefacts of shared aesthetic value indicates some shared concept of this value between people and cultures. Could machines ever share such concepts? This remains an open question, but evidence suggests that achieving such a unity would require consideration of factors beyond the quantifiable properties of objects themselves.

In this paper we have examined the relationship between complexity measures and personal or specific understandings of aesthetics. Our results suggest that some measures can serve as crude proxies for personal visual aesthetic judgement but the measure itself needs to be carefully selected. Complexity remains an enigmatic and contested player in the long-term game of computational aesthetics.

\begin{acknowledgements}
This research was supported by an Australian Research Council Grant, FT170100033. We would also like to thank Andy Lomas who worked with us on early parts of this research and provided one of the datasets used in this paper.
\end{acknowledgements}

%
%

\bibliographystyle{spmpsci}      

\bibliography{references}
\end{document}